\documentclass[runningheads]{llncs}
\usepackage{graphicx}
\usepackage{float} 
\usepackage{subfigure}
\usepackage{tabularx}
\usepackage{makecell}
\usepackage{multirow}  
\usepackage{booktabs}
\usepackage{enumitem}
\setlist{nolistsep}
\usepackage{hyperref}

\usepackage[longnamesfirst,square,numbers,comma,sort&compress]{natbib}
\begin{document}
\title{Supopulation-Specific Synthetic EHR for Better Mortality Prediction
%\thanks{Supported by organization x.}
}
%
%\titlerunning{Abbreviated paper title}
% If the paper title is too long for the running head, you can set
% an abbreviated paper title here
%
\author{Oriel Perets\orcidID{0000-0002-2079-6582} \and
Nadav Rappoport\orcidID{0000-0002-7218-2558}}
%
% First names are abbreviated in the running head.
% If there are more than two authors, 'et al.' is used.
%
\institute{Ben Gurion University, Beer-Sheva, Israel \\
\email{orielpe@post.bgu.ac.il}, \email{nadavrap@bgu.ac.il}\\
}
\maketitle              % typeset the header of the contribution
\begin{abstract}
Electronic Health Records (EHR) often exhibit biased representation across subpopulations (SPs) due to various factors such as demographics, clinical conditions, and the medical center type, this results in models that underperform for underrepresented SPs. To address this issue, we introduce a novel framework leveraging generative adversarial networks (GANs) to generate SP-specific synthetic data for augmenting the training dataset. Then, adopting an ensemble approach, training separate prediction models for each SP to boost performance and deliver customized predictions.

Our framework underwent testing on two datasets from the MIMIC database, demonstrating an improvement of 8\%-31\% in ROCAUC  for underrepresented SPs. We have made our code and ensemble models publicity available, contributing to advancing equitable healthcare predictive analytics.

\end{abstract}
\section{Background}
Electronic health records (EHRs) are valuable resources for healthcare research, providing a wide range of clinical data, including vital signs, demographics, diagnosis, admissions, and clinical notes \cite{jensen_mining_2012}. Previous research shows this data has the potential to be used to develop prediction and classification models \cite{hou_predicting_2020} enable the consolidation of risk score and alerting mechanisms \cite{churpek_internal_2020}, and inform treatment decisions. However, these records suffer from many challenges \cite{hripcsak_next-generation_2013}, including the underrepresentation of certain SPs, which can limit the generalizability of machine learning models trained on this data. Factors such as demographic distribution mostly due to the geographical location of the medical facility, disease prevalence, and medical facility type (e.g., children’s hospital, nursing home) can all contribute to the lack of representation in the data \cite{rahman_addressing_2013}. We address this issue by proposing an ensemble framework based on generative models. Our framework introduces a new solution for the problem of highly imbalanced medical datasets, using synthetic samples added to the training set of each SP and training an SP-specific prediction model on the augmented data.

Several existing approaches attempt to address the issue of underrepresentation in EHRs. Some approaches overcome the imbalance by rebalancing the data or re-weighting specific samples \cite{barda_addressing_2020}. Other approaches use synthetic data generation techniques \cite{kaur2019systematic}. Under-sampling includes using fewer samples from the majority population for rebalancing the training set. While over-sampling does the opposite. One popular approach for over-sampling is the Synthetic Minority Over-sampling Technique (SMOTE) \cite{chawla_smote_2002}, which creates synthetic samples for the minority class by interpolating between existing samples.

The contributions of this work are as follows:
\begin{enumerate}
    \item \textbf{GAN-based ensemble framework}. We propose an ensemble framework, for improving a model's prediction performance of underrepresented SPs in a dataset. We first identify poorly performing SP in the dataset, then iteratively generate and add varying amounts of synthetic samples to the training set of each SP separately. We then train an SP-specific prediction model using the augmented training data (comprised of both real and synthetic samples). We show our proposed framework offers better model performance for underrepresented SP. This framework is independent of the problem settings, including the prediction model type, the task at hand, and the feature(s) used to divide the samples into SPs.
    \item \textbf{An evaluation pipeline for synthetic data}. We designed and developed a pipeline for evaluating the effect of adding synthetic samples to the training set on model performance, the SP sample size, and the number of synthetic samples added. We implemented two real-world use cases and datasets to evaluate our framework. Our pipeline is generic in a way it can be used to evaluate the ML efficacy of any synthetic data generator.
\end{enumerate}

This approach can improve the generalizability of models trained on EHRs by countering SP underrepresentation. Our ensemble framework is a practical solution that can be applied to any EHR dataset to improve model performance on SPs. We compare our approach with several baselines, including the baseline ensemble with no dataset augmentation, and a sampling technique.

\section{Related Work}
The effects of imbalanced data distributions are well-posed in many applications \cite{kotsiantis_handling_2006}. The specific characteristics of EHRs make them more prone to highly imbalanced data, making their use for training machine learning models challenging \cite{hripcsak2013next}. Several approaches exist for coping with highly imbalanced data. Here, we refer to \texttt{class} as the feature, or a combination of features, used for categorizing the samples into SPs.

\subsection{Preprocessing Approaches}
Preprocessing approaches are data-centered approaches, which alter the training set itself, before model training. These are the simplest methods for dealing with imbalanced data. Under-sampling techniques remove samples from the majority class \cite{rahman_addressing_2013}. Chawala et al. \cite{chawla_smote_2002} introduced the Synthetic Minority Over-Sampling Technique (SMOTE), upon which variations of the approach have been suggested \cite{kovacs2019smote}.

\subsection{Algorithmic Approaches}
Algorithmic approaches do not alter the training data per se, but instead adjust the learning or decision process to alter the weight of samples from a given class in the dataset or modify a penalty per class, resulting in a form of cost-sensitive learning \cite{elkan2001foundations}. Barda et al. \cite{barda_addressing_2020} proposed a custom recalibration algorithm using PCE and FRAX to assess the calibration between SPs.

\subsection{Synthetic Data for Performance Boosting}
Synthetic data generation has been used to address several datasets' problems, including privacy and utility \cite{yoon_anonymization_2020, libbi_generating_2021}. To our knowledge, methods using synthetic data neither used an ensemble approach for SP-specific synthetic data generation nor evaluated the SP-specific prediction model performance. Other methods did not explore different amounts of synthetic samples and used an augmented dataset containing both real and synthetic samples to train the prediction models. These facts make it challenging to exploit synthetic data generation techniques to address the problem of imbalanced datasets and evaluate the efficacy of synthetic samples on SP-specific machine learning performance. To address this, we developed the proposed framework and evaluation pipeline.

\subsection{Relevant Techniques}
There are several research using ML efficacy i.e how well a model trained on synthetic data is compared to one trained using the real data, as a proxy for performance Che et al. \cite{che_boosting_2017} proposed ehrGAN, and SSL-GAN, a technique involving using a trained ehrGAN in a semi-supervised manner to boost a CNN-based prediction model performance. Their approach demonstrated better prediction model performance when training is solely based on synthetic samples. Chin-Cheong et al. \cite{chin-cheong_generation_2019} proposed an approach using WGAN for generating high-utility synthetic EHR to be used in place of real samples for training machine learning models. Their approach is evaluated for utility using binary classification trained on the synthetic samples generated and shows similar, but inferior performance to those trained on the real samples. Xu et al. \cite{xu_modeling_2019} proposed the Conditional Tabular GAN (CTGAN), which is used in this research.

\section{Method}
\subsection{Proposed method}
\begin{figure}
    \centering
    \includegraphics[width=1\textwidth]{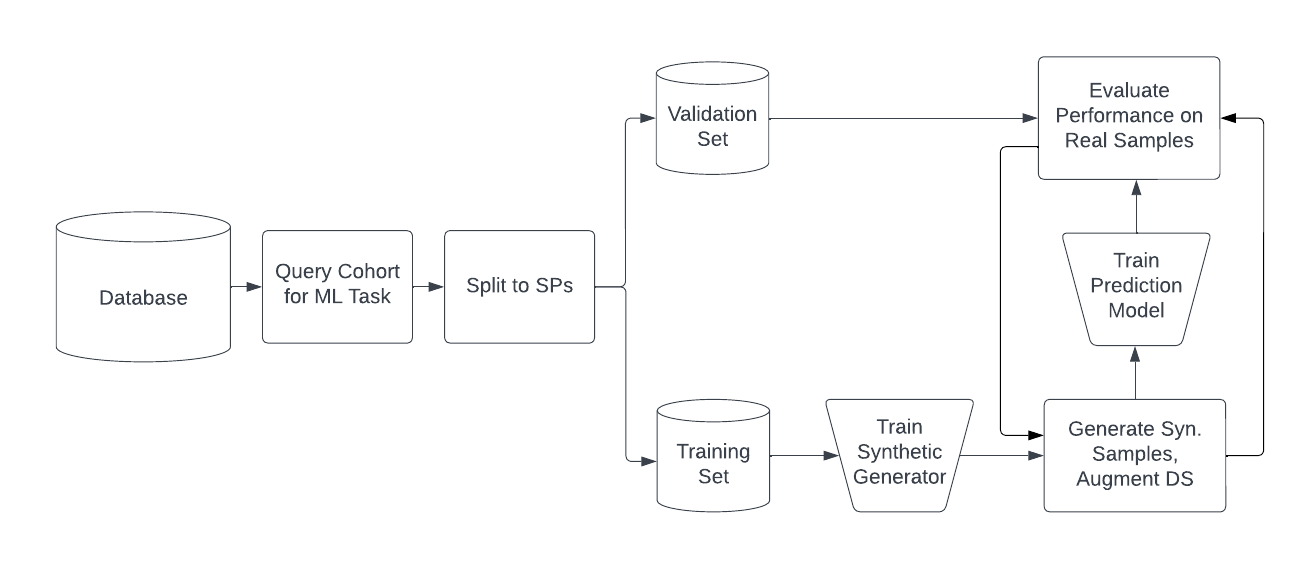}
    \caption{Method flow diagram; We query a cohort for the specific ML task, then split the dataset to SPs using the chosen PM, split each SP to training and validation sets, train the synthetic data generator, augment the dataset with synthetic samples, train the prediction model, and evaluate the performance using the validation set comprised of real samples. We then iterate the process for various amounts of synthetic samples.}
    \label{fig:spgan-flow}
\end{figure}

The proposed method is comprised of five steps, namely: (1) identifying under-performing SPs, (2) splitting the data into train and test sets, (3) training an SP-specific synthetic data generator, (4) generating and adding synthetic samples to the training set, and (5) evaluating the performance. A flowchart is presented in Figure \ref{fig:spgan-flow}.

\subsection{Datasets and Prediction Tasks}
We used two datasets and two corresponding prediction tasks. We used both MIMIC-III \cite{johnson2016mimic} and MIMIC-IV \cite{johnson2020mimic} databases to query the following datasets: (1) 30-day ICU mortality prediction for patients with sepsis-3 \cite{hou_predicting_2020} and (2) early mortality prediction for patients with CHF \cite{han_early_2022}. The datasets were queried by replicating the inclusion criteria detailed in the original papers. Data extraction queries are available in the supplementary material for ease of replication. Minimal preprocessing including dropping rows with missing values and label encoding was conducted.

\subsection{Subpopulation Definitions}
For each dataset at hand, we define `population marker` (PM) as the feature(s) used for the division into SPs. Here, we demonstrate such division using patients' ethnicity as the PM. Ethnicity was used for simplicity and popularity of the demographic feature in medical records and its known bias. However, our approach treats SP as an abstract concept and can be replicated using any feature or combination of features in the dataset, e.g., age group, lab results, vital signs, a combination of demographic features, etc. We define an SP suitable for applying this approach as one where model performance over it is systematically lower than the performance on the overall dataset.

\subsection{Ensemble-GAN Training}
Once patients are categorized into SPs, we (1) split the data into training (65\%) and test (35\%) sets, stratified by the outcome of interest to ensure enough samples from each outcome label are present for evaluating the prediction model and to ensure sufficient prevalence of the outcome label in the test set. We then use the training set to (2) train a Conditional Tabular Generative Adversarial Network (CTGAN) \cite{xu_modeling_2019} on each of the underrepresented SPs separately until convergence using an RTX6000 GPU. The model's parameters and the training time for each use case are detailed in the experiments section. (3) Generate synthetic samples in a range of amounts from 0 to 1000\% of the original training set size and add those samples to the training set of each SP.

\subsection{Prediction Models}
After acquiring the augmented training sets for each SP, we trained two prediction models, namely Gradient Boosting Classifiers (\hyperlink{https://scikit-learn.org/stable/modules/generated/sklearn.ensemble.GradientBoostingClassifier.html}{XGBoost}). These models were chosen for their popularity, performance, and ease of use. The models were implemented using the \hyperlink{https://scikit-learn.org/stable/}{Sklearn} library with set parameters to minimize variance between use cases and iterations. Two models were trained for each use case, one model is trained on the SP-specific augmented dataset, consisting of both the real samples (65\% \% of the SP samples) and the synthetic samples (0-1000\%, depending on the iteration). A second model is trained on 65\% of the full dataset, including all SPs, augmented with the SP-specific synthetic samples. This is done to evaluate the change in performance caused by the SP-specific synthetic samples over the entire dataset.

\begin{table}[H]
  \caption{Details of two prediction tasks. Number of samples, number of epochs, batch size, and training run-time for SP-GAN.}
  \label{tab:gan-training}
  \centering
  \begin{tabular}{llllll}
    \toprule
    Use case     & n   & Epochs     & Batch size   & Time (hrs)  \\
    \midrule
    Sepsis-3 & 4,559  & 300  & 100   & 02:12 \\
    CHF     & 11,062  & 300 & 100     & 04:46\\
    \bottomrule
  \end{tabular}
\end{table}

\subsection{Evaluation Pipeline}
Once the models are trained, we evaluate the models using ROCAUC. However, the evaluation pipeline allows using any metric, such as ROCAUC, accuracy, precision, recall, or PRAUC. The results are then compared for each percentage of synthetic samples added to the training set. Our released code for this evaluation pipeline can be used for any prediction task, dataset, and synthetic data generator. The code for this pipeline is available in the supplementary material.

\begin{figure}[H]
  \centering

  \subfigure[30-day mortality prediction for Sepsis-3 patients]{\includegraphics[width=0.4\textwidth]{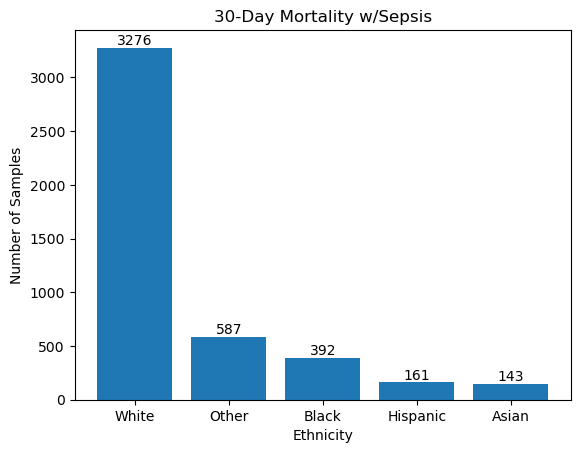}}
  \hspace{0.1\textwidth}
  \subfigure[Early mortality prediction for CHF patients]{\includegraphics[width=0.4\textwidth]{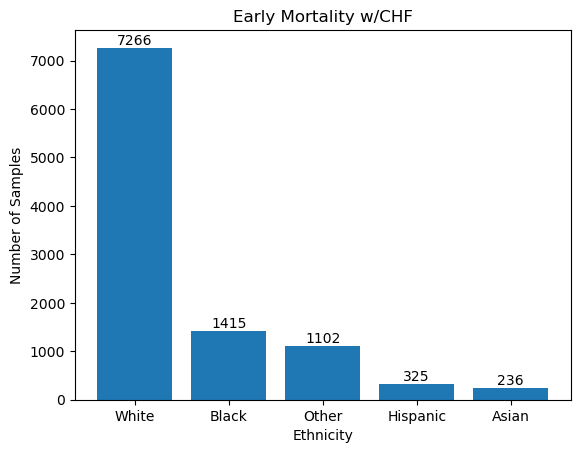}}

  \caption{SP sizes per use case ordered by size.}
  \label{fig:spsizes}
\end{figure}

\section{Experiments and Results}
We conducted experiments using the MIMIC III and MIMIC IV datasets for two use cases:

\begin{enumerate}[topsep=10px]
    \item \textbf{Early Mortality Prediction for Patients with CHF} This dataset contains 11,062 samples, each representing a CHF-diagnosed patient within the first 24 hours in the ICU.
    \item \textbf{30-day Mortality Prediction for Patients with Sepsis-3} This dataset comprises 4,559 samples, each representing a patient's first ICU admission with a Sepsis-3 diagnosis.
\end{enumerate}

We divided the datasets into SPs based on reported patient ethnicity: 'Black', 'Asian', 'Hispanic', 'White', and 'Other or Unknown'. SP sizes for each dataset are presented in Figure \ref{fig:spsizes}.

We further split the datasets into training (65\%) and testing (35\%) subsets while ensuring stratification by outcome label. This was done to guarantee adequate representation of each target class in the test set for model evaluation. We trained a baseline prediction model (\hyperlink{https://scikit-learn.org/stable/modules/generated/sklearn.ensemble.GradientBoostingClassifier.html}{XGBoost}) using real samples only from the training set for each SP, identifying underperforming SPs (ROCAUC lower than the full population).

'Hispanic', 'Asian', and 'Black' were identified as underperforming SPs for both use cases compared to the majority population ('White'). 'Other' was excluded from the experiment due to its lack of homogeneity, as an "Other" ethnicity can indicate a variety of patient ethnicities and encapsulate missing, unknown, or unable-to-obtain ethnicities.

After identifying underperforming SPs, we trained CTGAN models for each SP using only training set samples, without exposure to the test set. CTGAN models were trained with predefined hyperparameters (discriminator and generator learning rates, discriminator steps, epochs, and batch size) as outlined in Table \ref{tab:gan-training}. The trained CTGAN for each SP generated synthetic samples in varying amounts, from 0\% to 1000\% of the SP training set size. We evaluated the performance using ROCAUC for each SP and the percentage of added synthetic samples, results are presented in Table \ref{tab:comparisontable}. We compared the framework with two baselines:
\begin{enumerate}[topsep=10px]
    \item \textbf{Baseline ensemble learning} An individual model trained for each SP using only real samples.
    \item \textbf{Synthetic Minority Oversampling Technique (SMOTE)} Implemented using the \hyperlink{https://scikit-learn.org/stable/}{Sklearn} 'imblearn' library.
\end{enumerate}
For SMOTE, the data was split into train (65\%) and test (35\%) sets, stratified by ethnicity. SMOTE oversampling was performed on the training data, and an \hyperlink{https://scikit-learn.org/stable/modules/generated/sklearn.ensemble.GradientBoostingClassifier.html}{XGBoost} model was trained and evaluated for each SP. 
Ens. GAN model performance is ROCAUC for the optimal amount of synthetic samples added to the training set.
Our approach achieved better performance than the baseline and SMOTE models for 6 out of 6 under-represented SPs in two use cases (Table \ref{tab:comparisontable}).

% \begin{figure}[H]
%   \centering

%   \subfigure[30-day mortality prediction for Sepsis-3 patients]{\includegraphics[width=0.4\textwidth]{figures/sp_model_performance_30day.png}}
%   \hspace{0.1\textwidth}
%   \subfigure[Early mortality prediction for CHF patients]{\includegraphics[width=0.4\textwidth]{figures/sp_model_performance_chf.png}}

%   \caption{SP model performance as ROCAUC for each dataset}
%   \label{fig:spperformance}
% \end{figure}
\begin{table}[htbp]
  \centering
  \caption{Comparison of performance for each subpopulation, prediction task, and method. Results are presented as ROCAUC. n -  test set sample size. Ens. - Base ensemble model, trained on real SP samples and evaluated on real SP samples. Ens. GAN - our approach where models are trained on real plus synthetic samples per SP and tested on real SP samples. Ens. GAN's performance is presented from the optimal sample size.}
  \label{tab:comparisontable}
  \begin{tabular}{ccccccc}
    \toprule
    \textbf{Use Case} & \textbf{Subpopulation} &\textbf{n} & \textbf{SMOTE} & \textbf{Ens.} &\textbf{Ens. GAN}\\
    \midrule
    \multirow{4}{*}{\makecell{30 day mortality \\ with Sepsis-3}} & White & 3276  & - & 0.877 & 0.877 \\
                                & Other & 587  & - & 0.837 & 0.837\\
                                & Black & 392  & 0.796 & 0.805 & \textbf{0.834} \\
                                & Asian & 143 & 0.767  & 0.751 & \textbf{0.903} \\
                                & Hispanic & 161  & 0.642 & 0.556 & \textbf{0.695} \\
    \midrule
    \multirow{4}{*}{CHF mortality} & White & 7266  & -  & 0.723 & 0.725 \\
                                & Other & 1102  & - &  0.718 & 0.723\\
                                & Black & 1415  & 0.682 & 0.739 & \textbf{0.774} \\
                                & Asian & 325  & 0.579 &  0.673 & \textbf{0.721} \\
                                & Hispanic & 236 & 0.556 & 0.622 & \textbf{0.732}\\
    \bottomrule
  \end{tabular}
\end{table}

\section{Discussion}

In this paper, we attempt to address the issue of the underrepresentation of certain SPs in EHRs and their impact on the performance of ML models over SPs. We apply a SOTA synthetic data generation technique (CTGAN) \cite{xu_modeling_2019}. Together with an ensemble-like approach for generating and using synthetic samples. Inspired by our experimental results and based on two real-world use cases from the MIMIC database, we suggest our proposed ensemble framework improves the model's performance for SPs compared to other methods. Consequently increasing the generalizability of machine learning models towards SP that have a small amount of samples. This simple method is shown to increase the predictive power of ML models for those underrepresented SPs. Our framework consisted of training a GAN-based synthetic data generator, for each underrepresented SP separately, and incorporating the synthetic samples into the training set. Followed by training SP-specific prediction models using the augmented training data. Existing approaches either used synthetic samples to boost the full model performance or maintained their performance when trained on synthetic samples compared to real samples. While many datasets suffer from highly imbalanced classes and low representation of certain SPs, we emphasize our approach of adding synthetic samples to the real data, both broadly and specifically to improve SP-specific performance has not been explored. 

Furthermore, to evaluate the effectiveness of our method, we designed and implemented an evaluation pipeline using two real-world use case datasets queried from the MIMIC database. The evaluation pipeline evaluated both the efficacy of the synthetic samples, in terms of SP performance, and provided evidence our proposed framework has merit, compared to other approaches including SMOTE, and a vanilla ensemble. Our experimental results not only suggest better performance over data-balancing techniques but also preserve model performance over the entire population. The evaluation pipeline is available openly on \hyperlink{https://github.com/nadavlab/Improving_Subpopulation_Synthetic}{Github}

More research is needed to ensure the generalizability of the framework further and explore different amounts of synthetic data and other synthetic data generation methods, and different ways to define SPs. We used a categorical feature of race and ethnicity, but other data-driven approaches may be explored.
Our proposed method is well suited for federated learning where different SPs are separated and cannot be shared across sites.

A limitation of our suggested method compared to other methods is an increase in training time, as it requires training a GAN and generating samples, and training predictors using larger training sets. However, there is no effect on the running time during inference.

Our ensemble framework is a practical solution that can be applied to any EHR dataset to improve model performance and allow accurate predictions for small populations. It is independent of the prediction model, task, and population marker used for SP division, making it useful for other problems.

\bibliographystyle{splncs04}
\bibliography{main}
\end{document}